\documentclass[letterpaper, 10pt, conference]{ieeeconf}
\IEEEoverridecommandlockouts

\usepackage{cite}
\usepackage[utf8]{inputenc}
\usepackage{csquotes}
\usepackage{amsmath,amssymb,amsfonts}
\usepackage{balance}
\usepackage{dsfont}
\usepackage{censor}
\usepackage{algorithmic}
\usepackage{booktabs}
\usepackage{graphicx}
\usepackage[hidelinks]{hyperref}
 \usepackage{makecell}
\usepackage{textcomp}
\usepackage{xcolor}
\usepackage{tikz}
\usetikzlibrary{arrows.meta}
\usetikzlibrary{shapes.geometric} %
\usetikzlibrary{calc}

\usepackage[nolist,nohyperlinks]{acronym}

\begin{document}

\definecolor{markerspawn}{RGB}{0, 255, 255}  %
\definecolor{markergoal}{RGB}{0, 255, 0}      %
\definecolor{markerpath}{RGB}{255, 255, 0}    %

\DeclareRobustCommand{\msymspawn}{%
    \tikz[baseline=-0.5ex]\draw[fill=markerspawn, draw=black]
      (0,0) circle (0.70ex);\unskip%
}
\DeclareRobustCommand{\msymgoal}{%
    \tikz[baseline=-0.5ex]\node[star, star points=5, star point ratio=2.25,
      fill=markergoal, draw=black, line width=0.25pt,
      inner sep=0pt, minimum size=2.0ex] {};\unskip%
}
\DeclareRobustCommand{\msympath}{%
  \tikz[baseline=-0.4ex]{%
    \draw[black, line width=2.5pt] (0,0) -- (1.8em,0);%
    \draw[markerpath, line width=1.8pt] (0,0) -- (1.8em,0);%
    \filldraw[fill=markerpath, draw=black, line width=0.3pt] (0.9em,0) circle (2.0pt);
  }\unskip%
}

\definecolor{markerfall}{HTML}{E34948}
\DeclareRobustCommand{\msymfall}{%
  \tikz[baseline=-0.5ex]{%
    \draw[white, line width=2.0pt, line cap=round]
      (-0.45ex,-0.45ex) -- (0.45ex,0.45ex) (-0.45ex,0.45ex) -- (0.45ex,-0.45ex);%
    \draw[markerfall, line width=1.2pt, line cap=round]
      (-0.45ex,-0.45ex) -- (0.45ex,0.45ex) (-0.45ex,0.45ex) -- (0.45ex,-0.45ex);%
  }\unskip%
}

\tikzset{
  spawnmark/.style={circle, fill=markerspawn, draw=black, line width=0.6pt,
                    inner sep=0pt, minimum size=1.4ex},
  goalmark/.style={star, star points=5, star point ratio=2.25, fill=markergoal,
                   draw=black, line width=0.5pt, inner sep=0pt, minimum size=2.0ex},
}

\DeclareRobustCommand{\kf}[1]{
    \tikz[baseline=(c.base)]
    \node[circle, draw=black, fill=white, inner sep=0.5pt, minimum size=1.9ex, font=\scriptsize\bfseries] (c) {#1};
}

\newif\ifdraft

\setlength{\textfloatsep}{6pt plus 2pt minus 2pt}
\setlength{\dbltextfloatsep}{12pt plus 2pt minus 2pt}
\setlength{\floatsep}{6pt plus 2pt minus 2pt}
\setlength{\abovecaptionskip}{4pt}

\def\BibTeX{{\rm B\kern-.05em{\sc i\kern-.025em b}\kern-.08em
    T\kern-.1667em\lower.7ex\hbox{E}\kern-.125emX}}

    \title{LEAP: Learning Emergent Active Perception for Quadruped Navigation}

\iftrue
  \author{Bora G\"{o}kbakan$^{1}$, St\'{e}phane Caron$^{2}$, and Philippe Sou\`eres$^{3}$
        \thanks{*This work was supported by the PEPR O2R AS2 (No. ANR-22-EXOD-0006) through France 2030. %
        $^{1}$\textit{Inria} and \textit{DI ENS, PSL}, Paris, France, %
        $^{2}$\textit{Institute for Intelligent Systems and Robotics, CNRS} and \textit{Sorbonne University}, Paris, France, %
        $^{3}$\textit{LAAS-CNRS} and \textit{University of Toulouse}, Toulouse, France.}
    }
\else
    \author{
            Anonymous Authors
        }
\fi     \maketitle
    \begin{acronym}
        \acro{POMDP}{Partially Observable Markov Decision Process}
        \acro{PPO}{Proximal Policy Optimization}
        \acro{DDA}{Digital Differential Analyzer}
        \acro{ViT}{Vision Transformer}
        \acro{GT}{ground-truth}
        \acro{GAE}{Generalized Advantage Estimation}
    \end{acronym}

    \begin{abstract}
        Active perception allows autonomous agents to select their viewpoints rather than passively process the viewpoints given to them, enabling them to target where to reduce uncertainty about their environment. Learned systems typically encourage this behavior with hand-designed proxy objectives, such as coverage or curiosity bonuses, that may conflict with the task. In this work, we propose a method to learn \emph{emergent} active perception (LEAP) without augmentation of the task objective. We formulate the problem of goal-oriented navigation over hazardous terrains with goals that must be discovered visually. We then propose an architecture for navigation policies with active perception, and train them on a terrain curriculum where task pressure alone leads to the emergence of gaze control. Key to this emergence, LEAP works on a gaze-invariant representation that integrates depth images into egocentric belief maps. We validate its performance in held-out evaluation scenarios, where it achieves a $92.7\%$ success rate, compared to $74.2\%$ for scripted or $34.5\%$ for passive perception, and comes within $4.6$ points of a privileged oracle. We validate that LEAP navigation policies, unchanged, can be directly applied to steering quadrupedal locomotion policies in physics simulation.
    \end{abstract}

    \section{Introduction}
\begin{figure*}[t]
    \centering
    \begin{tikzpicture}
      \node[anchor=south west, inner sep=0] (img){
          \includegraphics[trim={0 14cm 0 0cm},clip,width=\linewidth]{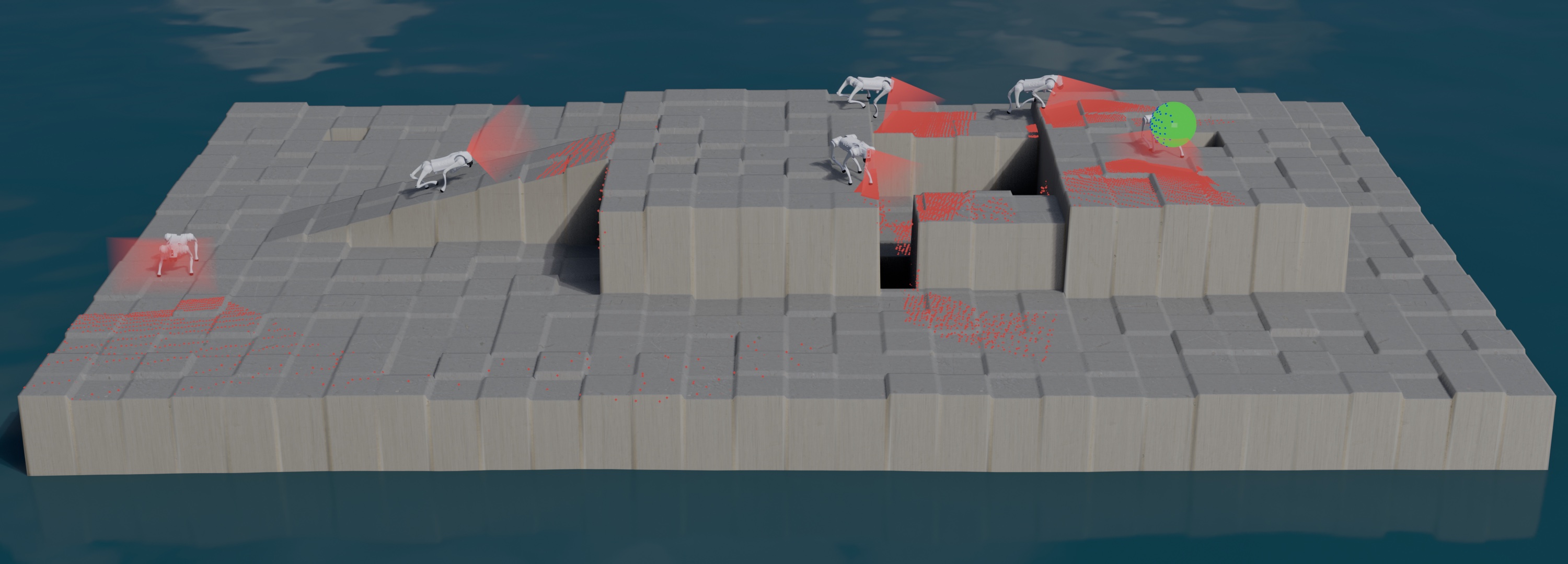}
      };
          \begin{scope}[x={(img.south east)}, y={(img.north west)}]

            \draw[-{Stealth[length=5pt]}, white, very thick] (0.20, 0.30) -- (0.24, 0.42);
            \draw[-{Stealth[length=5pt]}, white, very thick] (0.20, 0.30) -- (0.15, 0.30);                
            \node at (0.20, 0.30) {\kf{1}};                 %

            \draw[-{Stealth[length=5pt]}, white, very thick] (0.475, 0.75) -- (0.52, 0.75);
            \draw[-{Stealth[length=5pt]}, white, very thick] (0.475, 0.75) -- (0.52, 0.64);          
            \node at (0.475, 0.75) {\kf{2}};

            \draw[-{Stealth[length=5pt]}, white, very thick] (0.73, 0.93) -- (0.70, 0.835); 
            \draw[-{Stealth[length=5pt]}, white, very thick] (0.73, 0.93) -- (0.74, 0.75);
            \node at (0.725, 0.93) {\kf{3}};
          \end{scope}
        \end{tikzpicture}
        \caption{\textbf{Goal-oriented navigation on hazardous terrain.} 
        The Go2 quadruped must reach a goal whose location is not given, on a procedurally generated terrain, where the features that determine the route (a decoy crossing with a pit and a traversable bridge) are only visible from certain viewpoints. \textbf{LEAP} commands both the body and the camera, trained with no sensing reward. \kf{1}~At spawn, nothing is mapped: the robot must survey the surroundings before committing towards the platform; \kf{2}~the route choice hinges on which crossing is traversable: the robot must localize the hazard or rule it out before committing; \kf{3}~the robot must stay clear of  the final pit on the way to the goal. The rollout shown is the full-body sim2sim transfer on the \texttt{bridges} terrain.}
        \label{fig:teaser}

    \end{figure*}
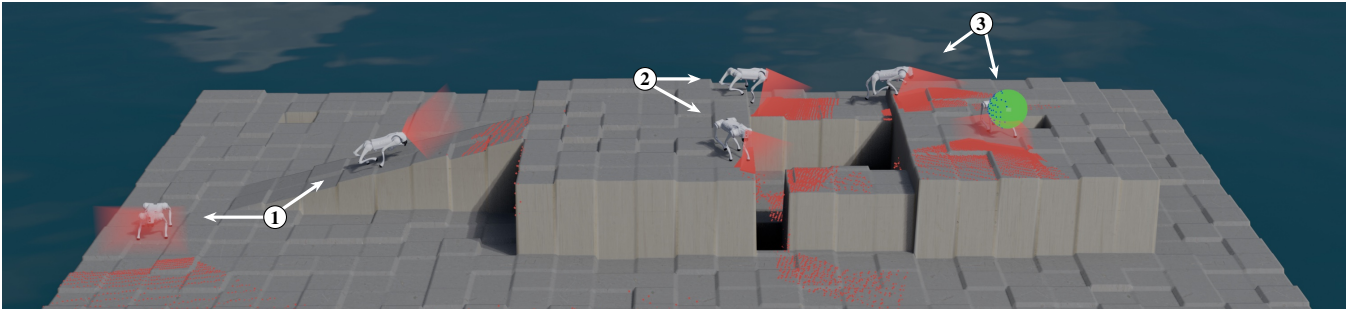

        Learned locomotion has enabled legged robots to perform agile movements and traverse obstacle courses, but most of its perception so far remains passive, with body-fixed cameras that undergo body movements without contributing to control decisions. This suffices where upcoming decisions enter the forward view successively~\cite{extreme-parkour,solo-parkour}, but there are more general situations where an agent will need to decide where to look before deciding where to go. Goal-oriented navigation on hazardous terrain (Fig.~\ref{fig:teaser}) is such a setting. The features that determine the feasibility of a route may be located below, beside, or behind the robot. A robot with a single limited-field-of-view sensor must then choose where to look.

        \looseness=-1 A glance that uncovers a hidden pit redirects the route; a glance at an already seen surface changes nothing. The value of a gaze is thus the improvement in expected return it enables. This \emph{value of information}~\cite{howard1966voi} is unobservable at decision time, so learned active-perception systems typically replace it with a hand-designed proxy: a curiosity bonus~\cite{pathak2017curiosity}, a coverage objective~\cite{chen2018coverage}, or a reconstruction loss~\cite{jayaraman2018learning}. The proxy is set independently of the task, so the two can disagree: an agent rewarded for coverage keeps scanning parts of the environment that are not relevant to the task; a robot with a direct line to the goal need not look behind. This raises the question of whether a separate sensing objective is needed at all. This paper proposes a framework for learning emergent active perception (\textbf{LEAP}), which removes the designed sensing objective altogether: a single policy controls both the body twist and the camera orientation, and with no sensing reward term, the return depends only on progress toward the goal. On the training terrains, a level forward gaze cannot resolve the hazards, and the goal must be discovered visually; an adaptive curriculum over spawn locations scales the difficulty as the policy improves. Under this pressure, gaze control emerges from the task reward alone.

        In summary, the contributions are: %
        \begin{itemize}
            \item a \ac{POMDP} formulation of goal-oriented navigation under active perception with no sensing reward, where gaze emerges from task pressure alone;
            \item a \emph{gaze-invariant} belief representation, where a cell is encoded identically whichever viewpoint observed it, so gaze exploration is not punished by a calibration error; a batched heightmap ray-caster renders depth at on-policy training scale;
            \item evidence that the emergent gaze is task-driven: $92.7\%$ held-out success over five terrain families, within $4.6$ points of a ground-truth oracle that, given the maps outright, never learns camera control.
        \end{itemize}
        The navigation policies trained with LEAP transfer to a full-body quadruped in physics simulation over held-out terrains, steering an independently trained locomotion policy.

    \section{Related Work}
        \subsection{Active Perception}\label{sec:active-perception}
            \looseness=-1 Most robot perception in learned locomotion is passive: the sensor position is fixed at design time, and perception is the problem of extracting the most from the input stream the placement yields. Active perception instead treats sensing as a control surface: where to point the sensor becomes a decision variable, made online, conditioned on the task~\cite{bajcsy1988,aloimonos1988active}. Learned implementations often make the sensing objective an \emph{explicit} target: \emph{information gain} motivates viewpoint selection in active SLAM~\cite{placed2023activeslam}, \emph{intrinsic curiosity} bonuses reward novelty during training~\cite{pathak2017curiosity}, and look-around policies are trained on \emph{coverage}~\cite{chen2018coverage} or \emph{scene reconstruction} objectives~\cite{jayaraman2018learning,ramakrishnan2019exploration}. Nevertheless, these surrogate terms can misprice the actual value of information: in tabletop manipulation, camera control learned from the task reward alone outperformed the same system trained with an added visibility bonus~\cite{cheng2018reinforcement}, and for a quadrotor learning goal-directed flight with an actuated camera, a voxel-based information-gain bonus added map coverage but no goal-reaching success~\cite{malczyk2026reinforcement}. LEAP takes the opposite position on the objective design: no reward term refers to the camera. Gaze control is learned only to the extent that it improves the task (\emph{i.e.}, navigation) return, so the information is priced by the task itself rather than approximated by a hand-designed heuristic. Closest to this position, \cite{kerr2025eyerobot} also learns camera control without a hand-designed sensing objective: a mechanical eyeball trained with RL is rewarded for views that help the manipulation policy it feeds imitate the demonstrations, and gaze behavior emerges from the coupling. LEAP differs in the task and in the coupling: goal-oriented navigation rather than tabletop manipulation, and a single policy commanding body and camera rather than separate perception (RL) and manipulation (imitation) learners.
            
        \subsection{Visual Legged Locomotion}\label{sec:visuo-loco}
            Advances in learning-based locomotion have allowed quadruped robots to traverse increasingly challenging terrains in recent years, progressing from blind locomotion on flat terrain~\cite{tan2018walking,hwangbo2019learning} to hiking on rough terrain, first blind~\cite{lee2020loco}, then with height scans sampled from elevation maps~\cite{miki2022perceptive}, and ultimately to parkour over obstacle courses with egocentric depth from body-fixed cameras alone~\cite{solo-parkour,extreme-parkour} or combined with LiDAR~\cite{hoeller2024parkour}. Sensing across this line of work remains passive: the sensors are body-fixed, and even the obstacle courses are linear, the terrain unrolling in front of the agent as it progresses. A forward-facing camera is sufficient to perceive the terrain ahead of the robot. LEAP drops this assumption: hazards often lie outside the body-fixed frustum in non-linear layouts, and viewpoint selection becomes a controller decision, with the same return steering both the body and the camera.

        \subsection{Viewpoint Calibration and Belief Maps}\label{sec:calibration_and_maps}
            Reinforcement learning directly from pixels is known to be sample-inefficient, and a sizable body of work facilitates it through the visual representation: contrastive auxiliary objectives~\cite{laskin2020curl} and image augmentation~\cite{yarats2022mastering} regularize the encoder against random crops and jitter. These augmentations are orthogonal to policy-driven viewpoint selection: a gaze change produces different content, and is not a mere perturbation of the percept. Encoder and policy are both calibrated to whatever viewpoint distribution the fixed sensor configuration produces. Active perception breaks that premise and closes a feedback loop: a novel viewpoint, however informative, falls outside the calibration, the return drops, and the viewpoint distribution never shifts. \emph{Moving the sensor hurts before it helps}. LEAP sidesteps the loop by construction: no learned encoder sits between the camera pose and the representation. Depth images are back-projected by fixed camera geometry into an egocentric bird's-eye-view belief map, a representation common in vision-based driving~\cite{ma2024bev}. The resulting representation is gaze-invariant: a new gaze changes \emph{which} cells are observed, but not \emph{how} observations are encoded.
            
            The map doubles as the policy's memory: instead of a recurrent state, the belief map persists what the gaze has observed. The belief map inherits from robotic elevation mapping, where range measurements are fused into height maps: \cite{fankhauser2018probabilistic} models the height uncertainty explicitly, and \cite{miki2022perceptive} couples such maps to locomotion, both under passive sensing. LEAP's map differs in how it is filled: the memory's contents are a policy decision.

    \section{Problem Statement}\label{sec:problem}
        In hazardous-terrain goal-oriented navigation, an agent spawns on terrain it has not seen before and must reach a goal whose location is not given. The features that decide the route (crossings, pits, dead ends) and the goal marker itself are visible only from the right viewpoints. The agent carries a single limited-field-of-view sensor whose orientation it controls. The problem is posed at the navigation level: the body is abstracted as a planar twist on the terrain surface, and the realization of the commanded twist by a legged platform is treated separately (Sec.~\ref{sec:method-locomotion-policy}).
        
        Formally, the navigation task is treated as a \ac{POMDP} $\mathcal{M} = \langle \mathcal{S}, \mathcal{A}, \mathcal{T}, \mathcal{O}, \mathcal{R}, \gamma \rangle $ where:
        \begin{itemize}
            \item $\mathcal{S}$ is the state space, where $\mathbf{s}_t = (\mathbf{q}_t, \boldsymbol{\eta}_t, \mathbf{x}^\text{goal}, m)$ collects the base pose $\mathbf{q}_t \in \mathrm{SE}(3)$, the sensor configuration (limited to its mechanical range), the goal position $\mathbf{x}^\text{goal} \in \mathbb{R}^3$, and the terrain $m$, fixed per episode. The height is bound to the terrain surface, here a heightfield $h$: $z_t = h(x_t, y_t)$ and roll/pitch align with the terrain normal, leaving position and yaw $(x_t, y_t, \psi_t)$ as the controllable degrees of freedom.
            \item $\mathcal{A}$ is the action space, where $\mathbf{a}_t = (\mathbf{a}_t^\text{body}, \mathbf{a}_t^\text{cam})$ pairs a planar body twist with the sensor's commanded velocity.
            \item $\mathcal{T}$ is the transition function: both actions are integrated under randomized actuation dynamics, accounting for the legged platform's imperfect tracking.
            \item $\mathcal{O}$ is the observation space: an exteroceptive measurement taken at the current sensor configuration $\boldsymbol{\eta}_t$, together with the robot state; the observation never includes $\mathbf{x}^\text{goal}$. The goal is observed only when the sensor points at it, so the agent must \emph{discover} the goal, and perception is active: through $\mathbf{a}_t^\text{cam}$, the agent controls what it sees.
            \item $\mathcal{R}: \mathcal{S} \times \mathcal{A} \times \mathcal{S} \to \mathbb{R}$ is the reward function. Sensing actions carry no direct reward: the reward depends only on the body position, never on $\boldsymbol{\eta}_t$ or $\mathbf{a}^\text{cam}_t$.
            \item $\gamma \in [0,1)$ is the discount factor.
        \end{itemize}
        
        Perception is instantiated with a depth camera on a pan-tilt mount: $\boldsymbol{\eta}_t \in \mathbb{R}^2$ are the pan-tilt angles, $\mathbf{a}_t^\text{cam} \in \mathbb{R}^2$ their commanded velocities, and the measurement is a depth image stacked with a binary channel encoding the goal. The formalism, however, applies to any agent-controlled sensor.

    \subsection{Cost-to-Go on Anisotropic Terrain}
\begin{figure}[t]
    \centering
    \includegraphics[trim={0 0 0.48cm 0},clip,width=0.75\columnwidth]{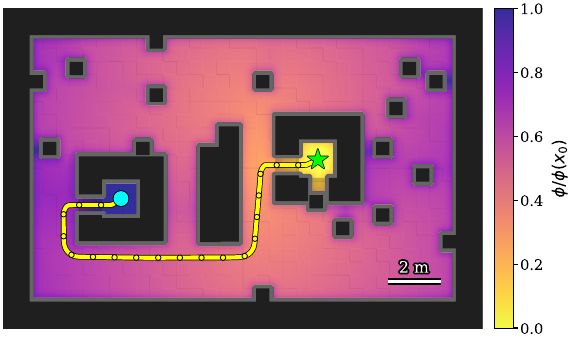}
    \caption{Normalized cost-to-go $\phi/\phi(\mathbf{x}_0^\ast)$ over an \texttt{islands} instance: the spawn and goal platforms are each ringed by a fatal moat crossed only by narrow crossings. The trace (\msympath) is the gradient descent on $\phi$ from the hardest spawn $\mathbf{x_0^\ast}$ (\msymspawn) to the goal $\mathbf{x^\mathrm{goal}}$  (\msymgoal); its nodes are the reverse-curriculum spawns. Black cores are the true obstacles; the lighter zone is the  agent-radius \emph{dilation} that the agent's center must avoid.}
    \label{fig:potential_map}
\end{figure}             %
            Progress toward the goal is measured by a cost-to-go $\phi(\mathbf{x})$: the cost of the cheapest path from the planar position $\mathbf{x}$ to the goal under a terrain-aware running cost, defined below. On flat ground far from hazards, $\phi$ reduces to the Euclidean distance to the goal.
            
            The cost-to-go is defined only over the geometrically feasible region, delimited by two constraints: (i) a maximum traversable height variation $\Delta h_\text{max} = 0.4$~m, the robot's standing body height, rules out both fall-height drops and unclimbable steps; (ii) the agent has a finite body width, disqualifying passages narrower than the body (Fig.~\ref{fig:potential_map}).  Infeasible states receive $\phi = +\infty$.
            
            Over the feasible region, the directional traversal cost combines a slope term and a clearance term:
            \begin{equation}
            c(\mathbf{x}, \hat{\mathbf{d}}) = 1 + w_h \bigl(\nabla h(\mathbf{x}) \cdot \hat{\mathbf{d}}\bigr)^2
                          + w_c \max\bigl(0, \, \delta - d_\text{c}(\mathbf{x})\bigr)^2,
            \end{equation}
            where $\nabla h(\mathbf{x})$ is the terrain slope at $\mathbf{x}$ and $\hat{\mathbf{d}}$ the direction of travel ($w_h = 0.5$), $d_\text{c}(\mathbf{x})$ is the clearance to the infeasible region, and $\delta = 0.575$~m is the cut-off distance. $w_c = 10$ discourages hugging the boundary of the infeasible region.
            
            Under the running cost $c$, $\phi$ satisfies the Bellman optimality condition
            \begin{equation}
                \min_{\hat{\mathbf{d}}\in\mathbb{S}^1}\bigl[c(\mathbf{x}, \hat{\mathbf{d}}) + \nabla\phi(\mathbf{x}) \cdot \hat{\mathbf{d}}\bigr] = 0,
                \qquad \phi(\mathbf{x}^\text{goal}_{xy}) = 0.
            \end{equation}

\begin{figure*}[t]
\centering
\begin{tikzpicture}
  \node[anchor=south west, inner sep=0] (img){
      \includegraphics[trim={0 0 0 0},clip,width=0.8\linewidth]{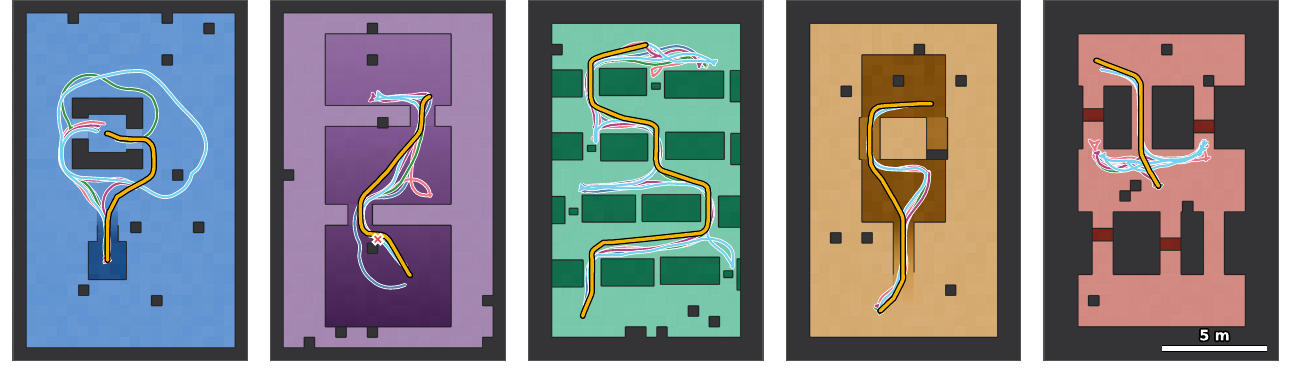}
  };
      \begin{scope}[x={(img.south east)}, y={(img.north west)}]

        \node[font=\ttfamily, anchor=south, yshift=2pt] at ($(img.north west)!0.10!(img.north east)$) {islands};
        \node[font=\ttfamily, anchor=south, yshift=2pt] at ($(img.north west)!0.30!(img.north east)$) {crests};
        \node[font=\ttfamily, anchor=south, yshift=2pt] at ($(img.north west)!0.50!(img.north east)$) {pillars};
        \node[font=\ttfamily, anchor=south, yshift=2pt] at ($(img.north west)!0.70!(img.north east)$) {bridges};
        \node[font=\ttfamily, anchor=south, yshift=2pt] at ($(img.north west)!0.90!(img.north east)$) {decoys};
        
        \node[spawnmark] at (0.083, 0.295) {};  \node[goalmark] at (0.083, 0.639) {};  %
        \node[spawnmark] at (0.333, 0.739) {};  \node[goalmark] at (0.318, 0.252) {};  %
        \node[spawnmark] at (0.500, 0.880) {};  \node[goalmark] at (0.452, 0.144) {};  %
        \node[spawnmark] at (0.681, 0.157) {};  \node[goalmark] at (0.722, 0.720) {};  %
        \node[spawnmark] at (0.897, 0.496) {};  \node[goalmark] at (0.848, 0.837) {};  %

      \end{scope}
    \end{tikzpicture}
    \caption{\textbf{Training environments.} One unseen instance of each procedural terrain type, with 25 \textbf{LEAP} rollouts (5 per tile); 24 reach the goal $\mathbf{x^\mathrm{goal}}$ (\msymgoal), \msymfall{} marks the single fall (\texttt{crests}), \msymspawn~the spawn location.}
    \label{fig:terrains}
\end{figure*}
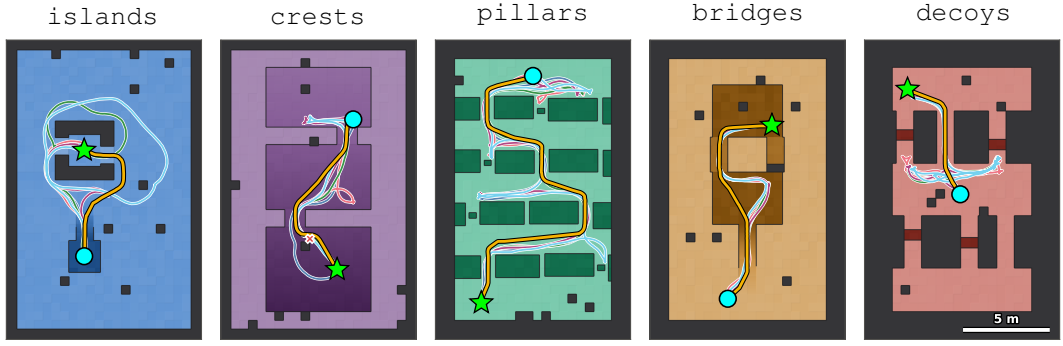
 
    \subsection{Reward}
        The reward is organized around the cost-to-go: $\phi$ supplies both the dense progress signal and the success criterion, and is complemented by a fall penalty and a constant time cost. The cost-to-go itself is computed from the ground-truth terrain and goal; the actor never observes it. The timestep reward
        \begin{equation}
            r_t = r^\text{progress}_t + r^\text{success}_t - r^\text{fall}_t - c_\tau
        \end{equation}
        combines dense progress shaping $r^\text{progress}_t$, defined below, with a sparse success bonus $r^\text{success}_t = \mathds{1}[\phi(\mathbf{x}_t) < r_\text{goal}]$, awarded when the cost-to-go drops below $r_\text{goal} = 0.5$~m, a fall penalty $r^\text{fall}_t = c_\text{f}\,\mathds{1}[\text{fall}_t]$ with $c_\text{f} = 2.5$, and a constant time cost $c_\tau = 0.002$.\footnote{Ablations showed that neither $c_\tau$ nor $r^\text{success}_t$ measurably affects the learned policy; both are retained as trained.} Success is measured geodesically rather than by planar distance, so a position behind an obstacle near the goal does not count as reaching it. A fall is a step across a drop exceeding $\Delta h_\text{max}$; a rise exceeding it blocks the motion instead, acting as a wall.\footnote{Since the fall position has $\phi = +\infty$, the shaping term is zeroed on that step: a fall is penalized by $r^\text{fall}$ alone.} Reaching the goal, falling, timing out, and leaving the terrain boundaries all terminate the episode.
        
        The shaping term $r^\text{progress}_t = \bigl(\phi(\mathbf{x}_{t-1}) - \phi(\mathbf{x}_t)\bigr)/\phi(\mathbf{x}_0^\ast)$, normalized by the cost-to-go at the nominal spawn $\mathbf{x}_0^\ast$ (the terrain's designated start position; the actual spawn $\mathbf{x}_0$ lies on the route from $\mathbf{x}_0^\ast$ to the goal $\mathbf{x}^\text{goal}_{xy}$), is potential-based reward shaping (cf.\ \cite{pbrs}): over any trajectory it telescopes to
        \begin{align}
            \phi(\mathbf{x}_0^\ast)^{-1} \sum_{t=1}^{T} \bigl(\phi(\mathbf{x}_{t-1}) - \phi(\mathbf{x}_t)\bigr) = \frac{\phi(\mathbf{x}_0) - \phi(\mathbf{x}_T)}{\phi(\mathbf{x}_0^\ast)},
        \end{align}
        independent of the path taken; since $\phi(\mathbf{x}_0) \le \phi(\mathbf{x}_0^\ast)$, a successful episode's cumulative progress reward is bounded by $1$. The telescoping has an important consequence for active perception: a probe-and-return detour nets exactly zero progress reward, and its only cost is the time it takes, through the discount $\gamma$ and $c_\tau$. The reward is therefore orthogonal to exploration: gaze earns nothing directly, and is rewarded only through the progress it enables toward the goal.

    \section{Method}
        The navigation policy $\pi^\text{nav}$ is trained in a custom GPU-parallel environment: procedurally generated hazardous terrains whose heightmaps and precomputed cost-to-go maps supply the reward. The observation pairs egocentric belief maps with the robot state, the action a unicycle twist with pan-tilt velocities; the policy is a Transformer encoder whose attention mask realizes an asymmetric actor-critic within a single network. The commanded twist is executed on the quadruped by a separate locomotion policy $\pi^\text{loco}$.

        \subsection{Terrain and Cost-to-Go Map}\label{sec:method-terrain}
            \paragraph{Terrains}
                Figure~\ref{fig:terrains} shows the five procedural families that are designed such that the hazards are not resolvable from a level, forward gaze. \texttt{islands} comprises elevated platforms and moat-ringed level platforms sampled around the spawn and goal, the moats crossable only via narrow bridges. \texttt{crests} is a slope whose three stages are connected by two depressed bridges, visible only up close. \texttt{pillars} is a dense field of pillars with traversable and dead-end alleys, making the crossing a search problem. \texttt{bridges} joins two decks by two bridges --- one traversable, one a decoy whose pit is occluded until the agent draws near. \texttt{decoys} spawns the agent on a center platform flanked by two outer platforms, each reachable by three short bridges of which exactly one is clear; the goal's side is randomized, so gaze must resolve both the side and the viable bridge. All tiles measure $16{\times}10$~m, additionally receive height jitter and random pit injection, and are discretized as heightmaps at $0.025$~m resolution.

            \paragraph{Geometric feasibility} 
                The two feasibility constraints from Section~\ref{sec:problem} are realized on the grid as follows: (i) cells where the local height range $h_\text{max}(\mathbf{x}) - h_\text{min}(\mathbf{x})$ over a one-cell neighborhood exceeds $\Delta h_\text{max}$ are marked untraversable; (ii) the agent is simulated as a point at its body center, and the untraversable mask is morphologically dilated by the footprint radius ($0.15$~m), sealing any passage narrower than the body width. The resulting feasible region is visible in Fig.~\ref{fig:potential_map}.

        \subsection{Observation Space}\label{sec:method-observation}
            At each step the agent receives a depth image from a body-mounted camera with controllable pan-tilt orientation $\boldsymbol{\eta}_t$ and fixed intrinsics modeling a RealSense D435.\footnote{$200{\times}116$~px, $87^\circ{\times}58^\circ$ FOV, $10$~m depth clip, zero-mean Gaussian range noise with $\sigma = 0.004\,z^2$, mounted $0.3$~m ahead and $0.4$~m above the body center with gimbal limits of $\pm85^\circ$ pan, $\pm60^\circ$ tilt.} Each image is integrated into two egocentric belief maps, a fine short-range map and a coarse long-range map (detailed below), which the policy observes together with the robot state. The maps are \emph{yaw-aligned}: each is a horizontal grid centered on the agent and rotated about the vertical axis to track the body heading; body pitch and roll do not tilt the map plane. Each cell exposes five channels $(\hat\mu, \hat\sigma, n_c, n_g, n_v)$: the mean and standard deviation of the surface height relative to the agent's foot level (in meters), a ray-hit count $n_c$ (the count of rays that update the cell), a goal-hit count $n_g$ (incremented by camera rays that intersect the goal marker: a floating sphere of $0.5$~m radius at a randomized height), and a visitation trail $n_v$ (tracing the agent's movement). Each depth image is transformed into a point cloud, and each point is accumulated into its cell's running height mean and variance. The goal location enters the observation only through $n_g$. Concretely, the observation includes:
    
            \begin{itemize}
                \item a \emph{fine} belief map $\mathbf{BM}_\text{fine}$ at $0.1$~m resolution ($\mathbb{R}^{5{\times}32{\times}32}$, covering $3.2{\times}3.2$~m around the agent); 
                \item a \emph{coarse} belief map $\mathbf{BM}_\text{coarse}$ at $0.6$~m resolution ($\mathbb{R}^{5{\times}32{\times}32}$, covering $19.2{\times}19.2$~m); 
                \item the robot state ($\mathbb{R}^{33}$):
                    \begin{equation*}
                        \begin{split}
                            \mathbf{s}_t^{\text{rob}} = \bigl(
                            &\mathbf{a}_{t-1}, v_t, \dot\psi_t,\, \ldots,\, \mathbf{a}_{t-4}, v_{t-3}, \dot\psi_{t-3},\\
                            &\hat{\mathbf{d}}^{\,\text{cam}}_t,\; \dot{\boldsymbol{\eta}}_t,\; z_t,\; \hat{\mathbf{g}}_t
                            \bigr),
                        \end{split}
                    \end{equation*}
                    a four-step window pairing each past raw action $\mathbf{a}_\tau$ (4) with the forward speed $v_{\tau+1}$ and yaw rate $\dot\psi_{\tau+1}$ it produced (making the per-episode actuation randomization identifiable), plus the body-frame camera direction $\mathbf{\hat{d}}^{\,\text{cam}}$ (3), the actual pan-tilt velocity $\dot{\boldsymbol{\eta}}$ (2), body height $z$ (1), and body-frame gravity $\hat{\mathbf{g}}$ (3).
            \end{itemize}
            Each cell's exposed height estimate is a posterior under a weak Gaussian prior of flat ground at foot level ($\sigma_0 = 1$\,m); an unobserved cell therefore reads $(0, 1.0\,\mathrm{m}, 0, 0, 0)$. Before each update, the evidence is decayed by a scale-dependent factor ($\alpha_\text{fine} = 0.93$, $\alpha_\text{coarse} = 0.99$), eroding $n_c$ toward zero and relaxing stale cells back toward the prior; one step of isotropic diffusion ($\lambda = 0.01$) then
  propagates the uncertainty spatially.
            
            \subsection{Action}
                At each timestep the navigation policy emits $\mathbf{a}_t = (\mathbf{a}_t^\text{body}, \mathbf{a}_t^\text{cam})$: a planar body twist and the pan-tilt velocities. The twist is restricted to a unicycle ($v_y \equiv 0$), leaving the forward speed ($v \in [-0.3, 1.5]$~m/s) and the yaw rate ($|\dot\psi| \leq 1.5$~rad/s), so the effective action space is $\mathbf{a}_t \in \mathbb{R}^4$. The unicycle model ties the body heading to the direction of travel, making active camera control the only way to look away from the heading. The commanded twist is smoothed by an exponential moving average on the body channels and integrated at 10\,Hz. The pan-tilt velocities ($\le 0.9$~rad/s per axis) are integrated and clipped to the gimbal limits.
    
        \subsection{Navigation Policy Architecture}\label{sec:method-policy}
            \begin{figure}[t]
                \centering
                \includegraphics[width=0.95\columnwidth]{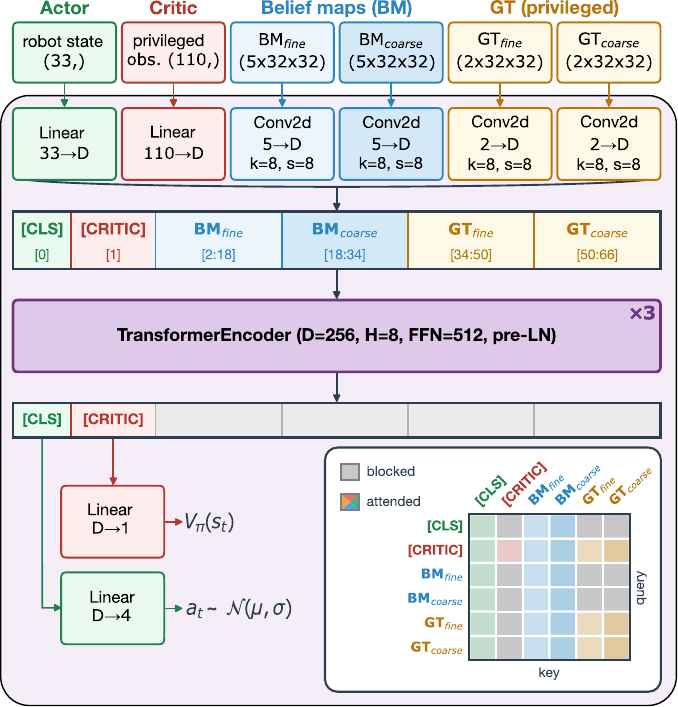}
                \caption{Policy architecture. The sparse attention mask allows a unified actor-critic network where privileged information is withheld from the actor.}
            \label{fig:policy}
            \end{figure}
            The navigation policy $\pi^\text{nav}$ maps the observation to the action $\mathbf{a}_t \in \mathbb{R}^4$ via a Transformer encoder~\cite{vaswani2017} over a sequence of $66$ spatial and task tokens, following the \ac{ViT} patch tokenization scheme~\cite{dosovitskiy2021an}. The architecture comprises three stages (Fig.~\ref{fig:policy}): tokenization of the maps and the robot state, a masked encoder shared by actor and critic, and linear decode heads.

            \paragraph{Tokenization}
                Each fine and coarse belief map is tokenized by a non-overlapping $8{\times}8$ convolutional projection into $16$ patches, yielding $32$ belief-map tokens in total. For each scale, a privileged two-channel \ac{GT} map at the same resolution and coverage as the corresponding belief map, encoding relative height around the agent and a binary goal-occupancy mask, is tokenized identically during training, adding another $32$ tokens. Learned positional embeddings are added to each patch token, while separate projection weights implicitly encode map type without dedicated embeddings.
                
                \looseness=-1 Two special tokens prepend the sequence. \texttt{[CLS]} is a linear projection of the robot state $\mathbf{s}_t^\text{rob}$ that the policy reads from. \texttt{[CRITIC]} is a linear projection of the privileged terrain vector ($\mathbb{R}^{110}$): terrain type ($5$) and instance one-hots ($100$), the spawn-difficulty $d_i$ (1), the normalized $\phi$ at $\mathbf{x}_t$ (1), and the goal position in the yaw-aligned body frame (3).\footnote{At deployment \acs{GT} and \texttt{[CRITIC]} tokens are zero-filled and filtered out by the attention mask.}
                
            \paragraph{Transformer encoder}
                A $3$-layer Transformer encoder processes the token sequence. A single forward pass serves both actor and critic: attention masking enforces the asymmetric actor-critic~\cite{asymmetric2017} split within the shared encoder. \texttt{[CLS]} and the belief-map patches are blocked from attending to \texttt{[CRITIC]} and to any \ac{GT} token, ensuring the actor receives no privileged information through any direct or multi-hop path. \texttt{[CRITIC]} is unmasked and attends to the full sequence, giving the value function access to terrain identity.
    
            \paragraph{Decode}
                The actor reads the \texttt{[CLS]} output through a linear head, emitting the mean of a diagonal Gaussian over $\mathbf{a}_t$. The log-standard-deviation is a learned vector independent of the observation. The critic reads the \texttt{[CRITIC]} output through a separate linear value head.
    
        \subsection{Locomotion Policy}\label{sec:method-locomotion-policy}
            \looseness=-1 The locomotion policy $\pi^\text{loco}$ emits joint-position targets at 50\,Hz for the Unitree Go2 quadruped, conditioned on the base state (linear and angular velocity, projected gravity), joint positions and velocities, the previous action, the planar twist command (randomly sampled during training, supplied at 10\,Hz by $\pi^\text{nav}$ at deployment), and a local height scan. The training terrains carry the same hazards as the navigation level (pits, planks, steps), and gait-shaping rewards are added to favor a trotting gait with sufficient foot clearance and stride length. The critic observes a rectangular 136-ray height scan ($1.6{\times}0.7$\,m at $0.1$\,m resolution); the actor only a 19-dimensional T-shaped reduction: a forward strip along the commanded path and a lateral bar under the base, the bar guarding against lateral drift over bridge edges. The T-shape confines $\pi^\text{loco}$ to terrain compliance along the unicycle corridor, leaving hazard avoidance to $\pi^\text{nav}$: in development runs where the actor observed the full scan, $\pi^\text{loco}$ became overly conservative, slowing or refusing twist commands near pits and cliff edges that $\pi^\text{nav}$ had already accounted for.
        
            During training, the scan is rendered from \ac{GT} terrain; at test time it is sampled from $\mathbf{BM}_\text{fine}$. Unobserved cells fall back to the lowest observed height in the scan (a pessimistic prior), and the scan reads ground-truth heights for cells within 1\,m of the spawn point, covering the camera's under-body blind spot; the navigation policy's belief is never given ground truth.

\begin{figure*}[t]
\centering
\begin{tikzpicture}
  \node[anchor=south west, inner sep=0] (img){
      \includegraphics[trim={0 0.34cm 0 0.79cm},clip,width=0.875\linewidth]{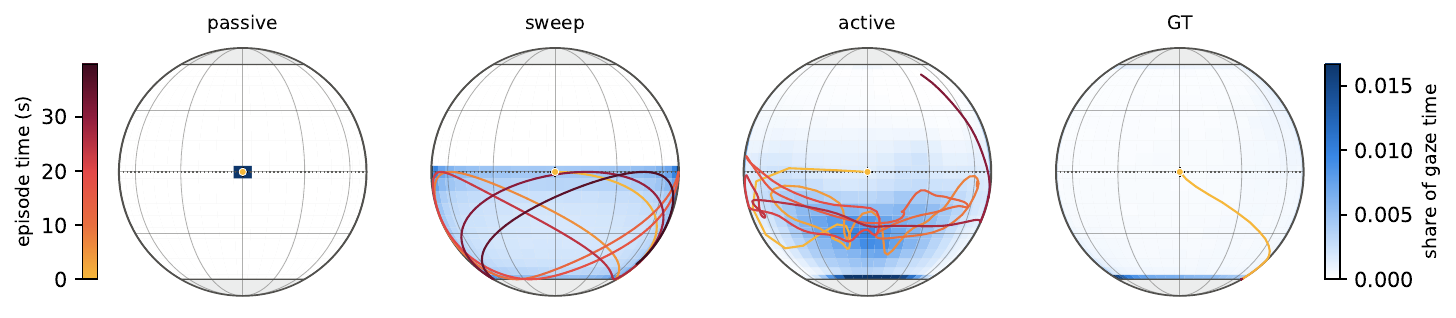}
  };
      \begin{scope}[x={(img.south east)}, y={(img.north west)}]
        \node[font=\upshape, anchor=south, yshift=2pt] at ($(img.north west)!0.168!(img.north east)$) {Passive};
        \node[font=\upshape, anchor=south, yshift=2pt] at ($(img.north west)!0.38!(img.north east)$) {Sweep};
        \node[font=\upshape, anchor=south, yshift=2pt] at ($(img.north west)!0.595!(img.north east)$) {Active};
        \node[font=\upshape, anchor=south, yshift=2pt] at ($(img.north west)!0.811!(img.north east)$) {GT};
      \end{scope}
    \end{tikzpicture}
    \caption{\textbf{Gaze allocation.} Camera direction distribution per variant (blue; $10\,000$ held-out episodes each, normalized to unit mass). The overlaid trace is a single episode on a shared \texttt{pillars} tile with the same initial spawn pose and neutral gaze $(0,0)$, colored by time. Passive never moves; Sweep follows its pre-programmed trajectory; Active (\textbf{LEAP}) concentrates gaze forward and down, along the direction of travel of the agent; \acs{GT} drifts into the corner and stops: with free information, looking just atrophies.}
    \label{fig:gaze_orbs}
\end{figure*}
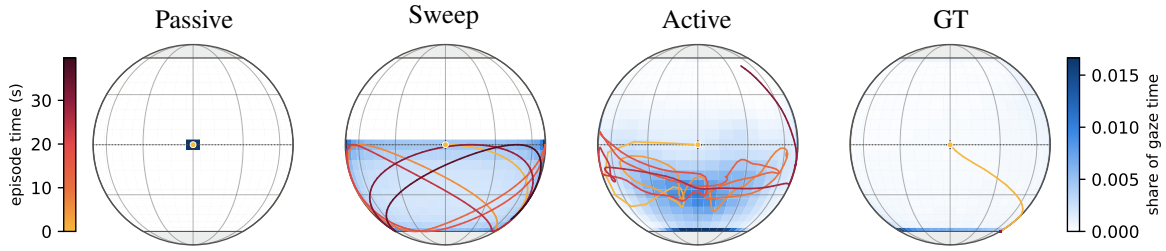
 
        \subsection{Training and Implementation}\label{sec:method-training}
            $\pi^\text{nav}$ is trained with \ac{PPO}~\cite{schulman2017} for $900\,\mathrm{M}$ steps\footnote{$400\,\mathrm{M}$ for \acs{GT} as it converges earlier (Sec.~\ref{sec:experiments}).} ($\approx 20$~h on an NVIDIA A100 GPU). Each episode spawns the agent at the curriculum-bound position $\mathbf{x}_0$ with heading $\psi_0$ sampled uniformly, and samples randomized actuation bounds: the maximum forward speed and yaw rate are scaled by independent factors uniform in $[0.5, 1]$, and the EMA smoothing coefficient $\alpha$ on the body action is drawn from $\mathcal{U}[0.15, 0.45]$: surrogates for the gain and tracking lag of the locomotion stage. The entropy coefficient is annealed from a positive (exploratory) value to a small negative one, shifting the policy from exploration toward a near-deterministic controller late in training. Advantages are estimated with \ac{GAE}~\cite{gae2016}. Remaining hyperparameters are listed in Table~\ref{tab:hparams}.

            \begin{table}[t]
                  \centering
                  \caption{Training hyperparameters.}
                  \label{tab:hparams}
                  \small
                  \setlength{\tabcolsep}{4pt}
                  \begin{tabular}{llll}
                      \toprule
                      Parallel envs     & $4096$          & Discount $\gamma$   & $0.99$ \\
                      Rollout length    & $64$            & \acs{GAE} $\lambda$ & $0.95$ \\
                      Minibatches       & $64$            & Optimizer & Adam \\
                      Update epochs     & $4$             & Value coef.\ $c_v$  & $0.5$ \\
                      Control step $dt$ & $0.1$\,s        & Grad-norm clip      & $0.2$ \\
                      Episode budget    & $40$--$150$\,s  & Curriculum step     & $\pm0.05$ \\
                      Learning rate       & $5{\times}10^{-4} \to 0$ (lin.)  & Clip $\epsilon$     & $0.2$ \\
                      Entropy coef.\ $c_e$ & $0.01 \to -0.005$ & & \\
                      \bottomrule
                  \end{tabular}
              \end{table}
    
            \paragraph{Spawn-distance Curriculum}
                \looseness=-1 Each terrain tile carries a waypoint chain, traced from its nominal starting point $\mathbf{x}_0^\ast$ to the goal via gradient descent on $\phi$. Each environment $i$ maintains a difficulty scalar $d_i$ that places the agent's spawn location $\mathbf{x}_0$ along this chain: $d_i = 1$ spawns at $\mathbf{x}_0^\ast$ itself, $d_i \approx 0$ near the goal. After each terminating episode, $d_i$ is updated independently per environment: it increases on success and decreases on a \emph{struggling} failure --- a fall or a timeout without sufficient progress (\emph{i.e.}, $\phi(\mathbf{x}_T) > 0.5\,\phi(\mathbf{x}_0)$). The episode time budget scales likewise with the spawn's cost-to-go, from $40$\,s near the goal to $150$\,s at the farthest nominal spawn.
                
            \paragraph{Cost-to-go solver}
                $\phi$ is solved on the heightmap grid by Jacobi iteration with a Godunov upwind scheme, implemented on the GPU with a custom Taichi kernel~\cite{taichi}.

            \paragraph{Depth rendering at scale}
                \looseness=-1 Each pixel of the depth image is ray-traced through the heightmap with a 2D \ac{DDA}~\cite{dda1987}, accelerated by two precomputed skip maps: a coarse max-pooled height grid for broadphase leaps above terrain, and an $L^2$ distance transform of occupancy that lets a ray leap the full distance to the nearest geometry. Both accelerations are conservative, so the output depth is \emph{exact} on the discrete heightmap. Implemented as a Taichi~\cite{taichi} kernel batched across all parallel environments, the renderer completes a full depth pass for $4096$ environments ($200{\times}116$ rays each) on a consumer GPU (RTX 4070~Ti~Super) at $11.5 \pm 3.0\times$ the throughput of the general-purpose IsaacLab ray-caster~\cite{isaaclab2025} on identical scenes (mean$\pm$std over the five terrain families). Substituted for the depth camera in the training pipeline of~\cite{extreme-parkour}, with the camera model and training configuration unchanged, it yields a $2.2\times$ end-to-end wall-clock speed-up: enabling training over thousands of on-policy environments with depth. The 2.5D heightmap precludes overhangs (none appear in training, as in~\cite{extreme-parkour}), and the renderer does not model self-occlusion by the robot body.

    \section{Experiments}\label{sec:experiments}
        \begin{table*}[t]
            \centering
            \caption{\textbf{Goal-reach success (\%).} mean$\pm$std over $5$ seeds; per seed, $4$ randomized episodes on each of $100$ unseen layouts per terrain type. Camera-control variants: Passive (fixed level gaze), Sweep (pre-programmed scanning), Active (learned gaze; \textbf{LEAP}), and \acs{GT} (oracle with ground-truth maps). \textbf{Bold:} best non-oracle success rate per terrain.}
            \label{tab:main}
            \setlength{\tabcolsep}{12pt}
            \begin{tabular}{lcccccc}
                \toprule
                & \texttt{islands} & \texttt{crests} & \texttt{pillars} & \texttt{bridges} & \texttt{decoys} & \emph{mean} \\
                \midrule
                Passive & $5.8 \pm 1.4$ & $2.5 \pm 1.9$ & $97.7 \pm 1.2$ & $36.6 \pm 2.8$ & $29.9 \pm 5.0$ & $34.5 \pm 1.6$ \\
                Sweep & $68.4 \pm 3.3$ & $79.1 \pm 2.2$ & $75.0 \pm 16.6$ & $85.6 \pm 2.0$ & $62.6 \pm 1.9$ & $74.2 \pm 3.3$ \\
                Active (\textbf{LEAP}) & $\mathbf{86.5 \pm 3.4}$ & $\mathbf{93.8 \pm 1.0}$ & $\mathbf{98.8 \pm 1.2}$ & $\mathbf{94.9 \pm 1.8}$ & $\mathbf{89.5 \pm 1.7}$ & $\mathbf{92.7 \pm 0.5}$ \\
                \midrule
                \acs{GT} & $94.9 \pm 0.8$ & $98.3 \pm 1.1$ & $99.5 \pm 0.3$ & $98.5 \pm 0.9$ & $95.3 \pm 1.1$ & $97.3 \pm 0.3$ \\
                \bottomrule
            \end{tabular}
        \end{table*}

        \begin{table}[t]
            \centering
            \caption{\textbf{Full-body transfer (MuJoCo).} Held-out success (\%, mean$\pm$std over 5 seeds); in parentheses, median time to goal (s) over successful episodes, averaged over seeds. \textbf{Bold:} best non-oracle success rate per terrain.
            }
            \label{tab:fullbody}
            \setlength{\tabcolsep}{2.75pt}
            \small
            \newcommand{\st}[2]{\makecell{$#1$\\[-2.5pt]{\scriptsize$(#2)$}}}
            \begin{tabular}{l ccc|c}
                \toprule
                & Passive & Sweep & Active (\textbf{LEAP}) & \acs{GT} \\
                \midrule
                \texttt{islands} & \st{18 \pm 6}{14.8} & \st{88 \pm 5}{33.9} & \st{\mathbf{97 \pm 2}}{16.2} & \st{91 \pm 4}{12.6} \\
                \addlinespace[2pt]
                \texttt{crests} & \st{16 \pm 15}{15.3} & \st{\mathbf{90 \pm 1}}{23.6} & \st{\mathbf{90 \pm 10}}{16.1} & \st{98 \pm 0}{11.7} \\
                \addlinespace[2pt]
                \texttt{pillars} & \st{79 \pm 9}{54.0} & \st{62 \pm 12}{106.4} & \st{\mathbf{88 \pm 15}}{33.3} & \st{86 \pm 7}{30.2} \\
                \addlinespace[2pt]
                \texttt{bridges} & \st{43 \pm 5}{12.1} & \st{80 \pm 6}{21.0} & \st{\mathbf{97 \pm 3}}{13.3} & \st{99 \pm 3}{10.7} \\
                \addlinespace[2pt]
                \texttt{decoys} & \st{36 \pm 7}{10.6} & \st{76 \pm 2}{22.0} & \st{\mathbf{87 \pm 6}}{12.1} & \st{96 \pm 3}{13.2} \\
                \midrule
                \emph{mean} & $38 \pm 5$ & $79 \pm 2$ & $\mathbf{92 \pm 6}$ & $94 \pm 2$ \\
                \bottomrule
            \end{tabular}
        \end{table}
        \looseness=-1 The experiments address three questions: (i)~does learned gaze control improve goal-reaching over passive and scripted camera sweeps (Sec.~\ref{sec:exp-main}); (ii)~\emph{what} gaze strategy emerges, and where blindness or indiscriminate scanning costs the alternatives (Sec.~\ref{sec:exp-gaze}); (iii)~does the policy transfer to a full-body quadruped in a robotics simulator with contacts (Sec.~\ref{sec:exp-fullbody})? Four camera-control variants are compared, defined in Table~\ref{tab:main}: \emph{Passive}, \emph{Sweep},\footnote{Sinusoidal sweeps: pan $\pm85^\circ$ ($12$\,s period); tilt $[-60^\circ, 0^\circ]$ ($\approx7.4$\,s). The periods are incommensurate and phases randomized per episode, so the pattern never repeats. Peak angular speed limited to the same $0.9$~rad/s as Active.} \emph{Active} (\textbf{LEAP}), and a privileged \acs{GT} oracle, an upper bound rather than a baseline.

        \subsection{Protocol}
            \looseness=-1 All four variants share the same architecture and observation space; Passive, Sweep, and Active also share the training budget, while \acs{GT} converges within a shorter one (Sec.~\ref{sec:method-training}). Passive, Sweep, and Active differ only in the camera control; the \acs{GT} variant differs from Active only in the attention mask, which additionally allows \texttt{[CLS]} and the \textbf{BM} tokens to attend to the \ac{GT} tokens: architecturally, \acs{GT} is a superset of Active. With no difference in learning capacity, the gap between the two isolates the cost of having to \emph{acquire} (partial) information. Each variant is trained with five random seeds. Evaluation uses held-out terrain: for each seed, $4$ randomized episodes on each of $100$ unseen layouts per terrain type ($2{,}000$ episodes per terrain--variant pair), with the agent spawned at the far end of the waypoint chain ($d=1$).
        
    \subsection{Goal-Reaching on Held-Out Terrain}\label{sec:exp-main}
        Table~\ref{tab:main} reports success on 100 unseen layouts. Learned gaze control outperforms the non-privileged variants: Active reaches $92.7\%$ mean success against $74.2\%$ for Sweep and $34.5\%$ for Passive, and comes within $4.6$ points of the \acs{GT} oracle. Active sensing recovers most of the task-relevant information a fully observable \ac{GT} map provides.
    
        The passive ablation exposes how terrain-dependent the value of gaze control is. On \texttt{pillars}, where the obstacles rise into a level forward view and the body-fixed camera pans with every turn, Passive matches Active ($97.7\%$ vs.\ $98.8\%$). Passive collapses on \texttt{crests} and \texttt{islands}: the crossings that make these terrains traversable lie \emph{below} a level gaze, and an agent that cannot look down cannot find them.
    
        The sweep ablation isolates the coverage factor. Sweep recovers much of what Passive misses, but lacks Active's \emph{timing}: looking at the bridge at the moment of crossing, or down at the ground while advancing over pits. In these scenarios, the ``forgetting'' of the belief maps (half-life of $\approx1$~s for $\mathbf{BM}_\text{fine}$) outpaces the indiscriminate gaze. A gaze schedule that ignores the search state wastes its time budget scanning task-irrelevant areas of the scene. One striking case is \texttt{pillars}, where sweeping degrades performance even below Passive ($75.0$ vs.\ $97.7$): Sweep pulls the gaze off the forward obstacles that a level view already resolves.
    
        \begin{figure}[t]
          \centering
          \includegraphics[width=0.8\columnwidth]{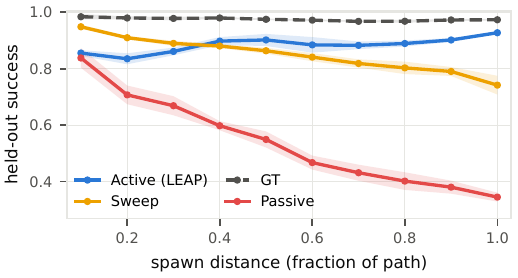}
          \caption{\textbf{Robustness to spawn distance.}
          Held-out terrain success versus spawn distance $d$: the fraction of the waypoint chain at which the agent spawns, equal to the difficulty scalar. %
          }
          \label{fig:robustness}
        \end{figure}
    
        \paragraph{Robustness to spawn distance}
            Fig.~\ref{fig:robustness} plots success as a function of spawn distance $d$. Active overtakes Sweep beyond $d\!\approx\!0.35$ and degrades more slowly. Active's weakness at small $d$ is a goal-discovery artifact rather than a navigation failure: training spawns lie far from the goal, so the policy learns to survey the terrain at episode start, and a near-goal spawn with a randomized goal-marker height demands a steeply upward gaze it rarely practices. Indeed, the fall rate at $d{=}0.1$ matches that at $d{=}1$ ($6.8\%$ vs.\ $6.9\%$), and the entire success gap is accounted for by the increased timeouts despite the goal proximity ($7.7\%$ vs.\ $0.3\%$).
        
        \subsection{Emergent Gaze Strategies}\label{sec:exp-gaze}
            A belief-conditioned gaze emerges only when the policy must rely on it. Fig.~\ref{fig:gaze_orbs} aggregates dwell-time statistics collected over $10{,}000$ episodes per variant, complemented with a traced instance of the camera trajectory. Active allocates the gaze asymmetrically: most of the time mass remains under the horizon with a forward pan, consistent with the agent's predominantly forward motion. The bulk of the camera movement is attributable to goal and terrain surveying. Sweep's gaze heatmap is symmetrical and broad: it is conditioned on neither the scene nor the completeness of the belief maps. Passive has a locked gaze: its successes concentrate on \texttt{pillars}, where the obstacles are visible at eye level and a fixed viewpoint suffices (Sec.~\ref{sec:exp-main}). 
            
            \looseness=-1 The \ac{GT} variant is supporting evidence that the \emph{emergent} active gaze is \emph{task-driven} rather than a random-saccade artifact of the action space. \ac{GT} shares the same camera control as Active, yet the gaze drifts and snaps to the gimbal-limit corner. Given privileged maps, the gaze becomes \emph{vestigial}. Camera control provides no utility and remains ``lazy''; it is learned \emph{when} and \emph{because} the policy relies on the belief it feeds.

        \subsection{Full-Body Sim2Sim Transfer}\label{sec:exp-fullbody}
            \looseness=-1 The planar training environment randomizes the actuation, and the body normal tracks the terrain normal, yet it abstracts leg contacts away: the robot cannot trip, slip, or lose its balance; ``falls'' are geometric rather than dynamic events. The navigation policy is transferred unchanged onto a full Go2 quadruped in MuJoCo~\cite{todorov2012mujoco}. The belief pipeline runs exactly as in training, and $\pi^\text{loco}$'s height scan is sampled from $\mathbf{BM}_\text{fine}$ rather than the simulator (except for \ac{GT}), so navigation and perception errors compound as they would on hardware. 
            
            An episode counts as a fall only on physically unrecoverable states (base tipped $>60^\circ$, or a ground-contact force on the base exceeding 25\% of body weight). Over 5 training seeds $\times$ 20 layouts $\times$ 6 episodes, all spawned at $d=1$ as in the main protocol, 600 episodes are collected per terrain and variant. Table~\ref{tab:fullbody} shows the ordering survives the transfer intact when averaged over terrains: \acs{GT} $\succ$ Active $\succ$ Sweep $\succ$ Passive. Absolute rates are not comparable to Table~\ref{tab:main}: the embodiments are different, the pool of held-out tiles and trials is smaller, and the geometric ``falls'' that terminate a planar episode are here often recoverable stumbles.

        \section{Conclusion}
            \looseness=-1 This paper introduced LEAP, a method to learn emergent active perception where gaze control emerges from the task rather than from custom, task-competing sensing rewards. The policy learns where to look, locates goals, and avoids hazards in terrains where passive perception fails. The ability to train successfully hinges on gaze-invariant belief-map representations and a suitable terrain curriculum. In evaluation over held-out terrains, LEAP reaches a $92.7\%$ success rate, against $74.2\%$ for scripted camera sweeps and $34.5\%$ for passive perception. As is, LEAP navigation policies can steer separately trained locomotion policies on a full-body quadruped in simulation.
            
            \looseness=-1 The observation design underlies the results. LEAP fuses camera images into gaze-invariant belief maps, so no reachable viewpoint is out of distribution and gaze exploration escapes the calibration penalty. The belief maps also act as a local spatio-temporal memory, sparing the network a recurrent state, but the 2.5D representation cannot capture overhangs. While the interface between navigation and locomotion is currently frozen, with the two policies trained independently, future work could investigate the benefits of training them jointly. LEAP could also be extended to humanoid robots, where the body twist is commanded at the pelvis and the camera sits higher at the head. Their ability to negotiate confined spaces will likely surface new tradeoffs between perception and body motion.

    \bibliographystyle{IEEEtran}
\balance

\end{document}